%% file: emnlp2021.tex
\pdfoutput=1

\documentclass[11pt]{article}

\usepackage{xr}
\usepackage[]{emnlp2021}

\usepackage{times}
\usepackage{latexsym}

\usepackage[T1]{fontenc}

\usepackage[utf8]{inputenc}

\usepackage{microtype}

\newcommand{\isection}[2]{\section{#1}\label{ssec:#2}}

\newcommand{\isubsection}[2]{\subsection{#1}\label{ssec:#2}}

\newcommand{\ex}[1]{``\emph{#1}''}

\usepackage{microtype}

\usepackage[inline]{enumitem}
\usepackage{graphicx}
\usepackage{multirow, booktabs}
\usepackage{soul}

\newcommand{\wow}[0]{{\sc WoW}}

\usepackage{graphics}
\usepackage{subcaption}

\makeatletter
\newcommand*{\addFileDependency}[1]{
  \typeout{(#1)}
  \@addtofilelist{#1}
  \IfFileExists{#1}{}{\typeout{No file #1.}}
}
\makeatother

\newcommand*{\myexternaldocument}[1]{%
    \externaldocument{#1}%
    \addFileDependency{#1.tex}%
    \addFileDependency{#1.aux}%
}

\myexternaldocument{appendix}

\title{$Q^{2}$: Evaluating Factual Consistency in Knowledge-Grounded Dialogues via Question Generation and Question Answering}

\author{
Or Honovich$^{1}$\thanks{\hspace{5px}Work done during an internship at Google Research.} \quad
Leshem Choshen$^{1}$ \quad
Roee Aharoni$^{2}$ \quad
\textbf{Ella Neeman}$^{1}$ \\
\textbf{Idan Szpektor}$^{2}$ \quad
\textbf{Omri Abend}$^{1}$ \\
$^1$The Hebrew University of Jerusalem;
$^2$Google Research\\
{\tt or.honovich@gmail.com} \\ {\tt \{roeeaharoni,szpektor\}@google.com}
}
\begin{document}
\maketitle
\begin{abstract}

Neural knowledge-grounded generative models for dialogue often produce content that is \textit{factually inconsistent} with the knowledge they rely on, making them unreliable and limiting their applicability. 
Inspired by recent work on evaluating factual consistency in abstractive summarization, 
we propose an automatic evaluation metric for factual consistency in knowledge-grounded dialogue using automatic question generation and question answering. Our metric, denoted $Q^2$, compares answer spans using natural language inference (NLI), instead of token-based matching as done in previous work.
To foster proper evaluation, we curate a novel dataset of dialogue system outputs for the Wizard-of-Wikipedia dataset, manually annotated for factual consistency.
We perform a thorough meta-evaluation of $Q^2$ against other metrics using this dataset and two others, where it consistently shows higher correlation with human judgements.

\end{abstract}

\isection{Introduction}{sec:introduction}

\begin{figure}[t]
\centering
 \includegraphics[width=1.0\linewidth]{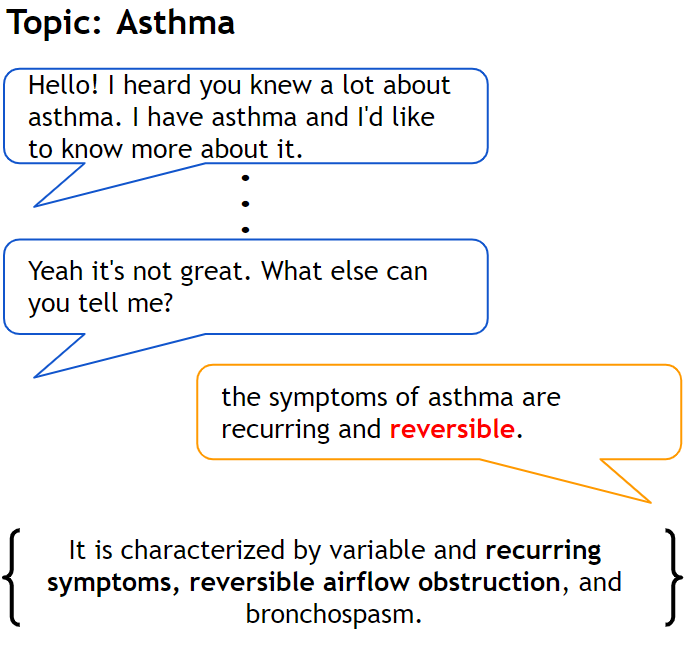}
 \caption{An example from our dataset. Human messages are in Blue, the generated response is in Orange and the grounding knowledge is in Black at the bottom. The factual inconsistency is marked in Red.}
 \label{fig:intro_snippet}
\end{figure}

Generative conversational agents show remarkable progress lately \citep{shuster-etal-2020-dialogue, adiwardana2020towards}.
Yet, generative dialogue models that are grounded by external knowledge sources
still struggle to be consistent with that knowledge. Their output is often incompatible with the given knowledge or even completely ``hallucinated'' \citep{roller2020recipes}.
Figure~\ref{fig:intro_snippet} depicts such inconsistency by the dialogue system of \citet{shuster-etal-2020-dialogue} when evaluated on the {W}izard of {W}ikipedia dataset \citep{dinan2019wizard}. 
Since inconsistent generated text is usually fluent and well-formed, these outputs could mislead users with false information, limiting the applicability of such systems.

Factual inconsistency is often overlooked by evaluation methods for text generation \citep{celikyilmaz2020evaluation}. 
Evaluation approaches that address this gap were recently proposed for tasks like machine translation and abstractive summarization \citep{sellam-etal-2020-bleurt, xu-etal-2020-fact, goodrich2019assessing}. Yet, evaluating grounded dialogues poses additional challenges, 
since dialogue outputs may refer to the dialogue history and include personal opinions, questions to the user, and general ``chit-chat'', whose consistency with external knowledge is mostly irrelevant. Additionally, many of those metrics require gold-label human-constructed references, while dialogue is an open-ended task -- making it less suitable for reference-based evaluation. 


In this work, we propose an automatic metric for evaluating the factual consistency of generative open-domain knowledge-grounded dialogue systems which does not require gold-label reference responses. 
Our metric, denoted $Q^2$, pairs automatic question generation (QG) and question answering (QA) for dialogue generation evaluation, inspired by recent work on factual consistency evaluation in abstractive summarization \citep{durmus-etal-2020-feqa, wang-etal-2020-asking}. 
$Q^2$ first takes a given generated response as input, and generates questions whose answers are informative spans in the response, using a QG system. 
It then employs a QA system to find corresponding answer spans in the knowledge that the response should be grounded in. The evaluation score reflects the similarity between each informative response span and its corresponding answer span from the knowledge, for each generated question.

Unlike previous QG/QA approaches, which used token-based matching to compare answer spans, we propose a novel comparison method using natural language inference models \citep[NLI;][]{dagan-pascal-2003}  that is more robust to lexical variability. In addition, while QG/QA based methods showed promising results for summarization evaluation, our work is the first to apply them to knowledge-grounded dialogues, which hold distinct properties compared to other grounded generation tasks; Mixing different types of utterances such as knowledge, personal statements and chit-chat in a single response is unique to dialogue and is well addressed by our metric given its modular nature and robustness to lexical variability.

We assess $Q^2$ against other reference-response-free metrics on three dialogue benchmarks: {W}izard of {W}ikipedia \citep[\wow{};][] {dinan2019wizard}, Topical-Chat \citep{Gopalakrishnan2019} and Dialogue NLI \citep[DNLI;][]{welleck-etal-2019-dialogue}. To foster proper evaluation, we curate a new dataset of dialogue system responses using the \wow{} dataset, manually annotated for factual consistency. 
$Q^2$ reaches significantly higher correlations with human judgments on all datasets compared to the other metrics, demonstrating its potential as an evaluation framework for grounded dialogue generation.

To summarize, our contributions in this work are three-fold: \begin{enumerate*}[label=(\arabic*)]
  \item We develop a novel framework for evaluating the factual consistency of knowledge-grounded, open-domain dialogue systems, incorporating question generation, question answering and NLI models.
  \item We construct a first-of-its-kind dataset of knowledge-grounded dialogue system outputs manually annotated for factual consistency, fostering future work on the subject.
  \item We validate the effectiveness of our metric in comparison to previous approaches through various experiments with three dialogue benchmarks, where it obtains higher correlation with human judgements.\vspace{-5px}\footnote{Our code and dataset are available in: \url{http://github.com/orhonovich/q-squared}}
\end{enumerate*}

\begin{figure}[t]
\centering
 \includegraphics[width=0.485\textwidth]{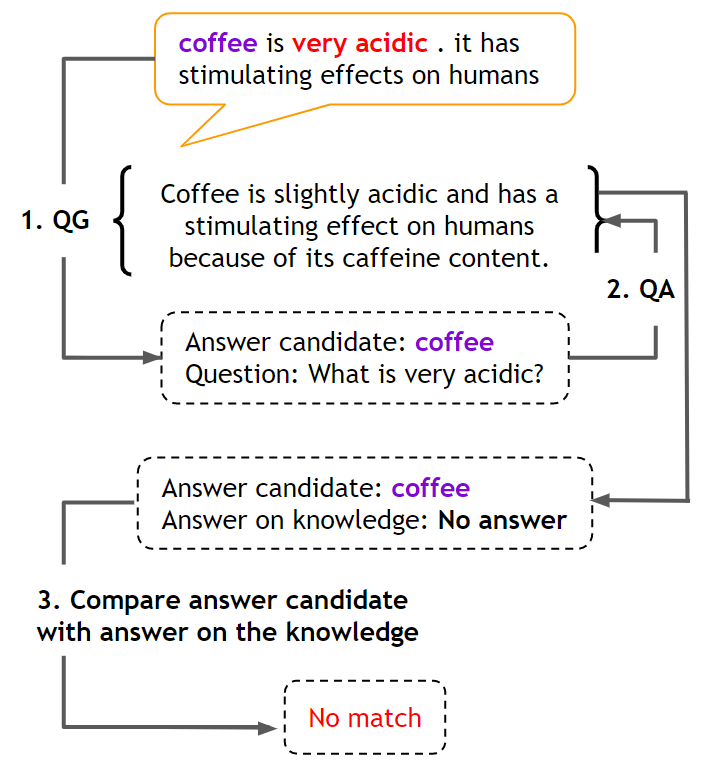}
 \caption{The $Q^{2}$ pipeline: 
 \begin{enumerate*}[label=(\arabic*)]
  \item For a response, select answer candidates; then generate a question for each candidate using QG. 
  \item Use QA to answer each question based on the grounding knowledge. 
  \item Compare the answer candidate with the knowledge answer span.
 \end{enumerate*}}
 \label{fig:pipeline}
\end{figure}
\section{Evaluating Factual Consistency}
\label{sec:pipeline}

Formally, an evaluation metric for factual consistency in generative dialogue receives as input a dialogue history $h$, a textual knowledge source $k$, and a response $r$ from a dialogue model (assumed to be generated conditioning on $h$ and $k$). The goal is to score the model's output $r$ so as to reflect its consistency with its grounding source $k$.
We next introduce our metric, denoted $Q^2$,
which suggests that factual questions that have answers in the generated response should have similar answers in the grounding knowledge source, while differences between answers from the response and the knowledge point at factual inconsistencies. This follows the intuition in \citet{wang-etal-2020-asking, durmus-etal-2020-feqa} for evaluating abstractive summarization.

$Q^2$ iterates over all informative spans $a^r_i$ in $r$. For each $a^r_i$, $Q^2$ uses a QG system to generate questions $q_{i_j}$ whose answer is $a^r_i$. For each question $q_{i_j}$, $Q^2$ uses an extractive QA system to mark an answer span $a^k_{i_j}$ from $k$. $Q^2$ then measures the similarity of $a^r_i$ and $a^k_{i_j}$ and aggregates the similarity scores for all questions as the factual consistency score of $r$.
Figure \ref{fig:pipeline} depicts this procedure. We next detail each component in our metric.




\paragraph{Question Generation.}
First, we mark informative spans in the response $r$ to serve as target answer spans for the QG system. To this end, we mark all named entities and noun phrases in $r$ using spaCy.\footnote{\url{https://spacy.io/}} For example, in ``\emph{coffee is very acidic}'' we mark `\emph{coffee}’  as an informative span.
Then, a QG system takes each informative span $a^r_i$ and the response $r$ as input and generates the corresponding questions $q_{i_j}$ for which $a^r_i$ should be the answer. In our example, a generated question for the informative span `\textit{coffee}' and the response in Figure \ref{fig:pipeline} is ``\textit{What is very acidic?}''. We use T5-base \citep{2020t5} fine-tuned on SQuAD1.1 \citep{rajpurkar-etal-2016-squad} as the QG model.\footnote{\url{https://huggingface.co/mrm8488/t5-base-finetuned-question-generation-ap}}

As suggested by \citet{wang-etal-2020-asking}, we use beam search decoding, taking the top-n generated questions for $a^r_i$. We set $n = 5$ and test two variants of generating multiple questions. In the first, we use all $n$ questions for $a^r_i$. In the second variant, we only take the top-ranked question that passed the filtering stage for $a^r_i$ (see ``Question Filtering'' below). 
We observed similar trends for both variants, and therefore only report the results of the second variant. To increase the diversity of the generated questions, we tried sampling-based methods \citep{fan-etal-2018-hierarchical, Holtzman2020degeneration}, but obtained inferior results that are not reported in this paper.

\paragraph{Question Answering.}
To mark the answer span $a^k_{i_j}$ in the knowledge $k$ for question $q_{i_j}$, we use the Albert-Xlarge model \citep{Lan2020ALBERT} fine-tuned on SQuAD2.0 \citep{rajpurkar-etal-2018-know}.\footnote{\url{https://huggingface.co/ktrapeznikov/albert-xlarge-v2-squad-v2}}
This model can also determine that no answer can be found in the  paragraph. This is important in $Q^2$, since question $q_{i_j}$ generated for a completely hallucinated content $a^r_i$ should have no answer in $k$.


\paragraph{Answer Similarity and Final Scores.}


The last step in $Q^2$ assesses the similarity between answers $a^r_i$ and $a^k_{i_j}$. To be robust to lexical variability between the response and the knowledge, e.g. \ex{US} vs. \ex{United States} or \ex{a book series} vs. \ex{a set of novels}, we measure the answer span similarity using an NLI model. We use RoBERTa \citep{liu2019roberta} fine-tuned on SNLI \citep{bowman-etal-2015-large} as implemented in AllenNLP ~\citep{Gardner2017AllenNLP}.

For span pairs $a^r_i$ and $a^k_{i_j}$ that match perfectly at the token-level, we assign a score of $1$. For each span pair $a^r_i$ and $a^k_{i_j}$ that do not match perfectly at the token-level, 
we run the NLI model with $a^k_{i_j}$ as the premise and $a^r_i$ as the hypothesis. To add context for the NLI model, each answer is concatenated after the question $q_{i_j}$.
For example, for the question \ex{Where were the Red Hot Chili Peppers formed?}, the response answer \ex{LA}, and the knowledge answer \ex{Los Angeles}, we run the NLI model with: \ex{Where were the Red Hot Chili Peppers formed? Los Angeles} as the premise, and with \ex{Where were the Red Hot Chili Peppers formed? LA} as the hypothesis. Our use of NLI differs from prior use of NLI in dialogue evaluation, where it was applied in an end-to-end manner \cite{welleck-etal-2019-dialogue,pang-etal-2020-towards}.
We set $q_{i_j}$'s score to be $1$ for the case of entailment and $0$ for contradiction or for cases where the QA model produced no answer. In the neutral case, we take the answers token-level F1 score, as in \citet{wang-etal-2020-asking}.

Finally, the match scores for all answer pairs are averaged to yield a response-level score, and the response-level scores are averaged to yield a system-level $Q^{2}$ score.

\paragraph{Question Filtering.}

To alleviate errors made by the automatic QG and QA models, we follow the validation step in \citet{wang-etal-2020-asking}; We run the QA model to answer $q_{i_j}$ with the response $r$ as the input paragraph, and require the answer to be identical to the answer span $a^r_i$ which was used to generate $q_{i_j}$. If this is not the case, $q_{i_j}$ is discarded. 

As we evaluate factual consistency, we wish to ignore opinionated parts of the response which are not factual. Hence, we filter out questions that include the personal pronouns \ex{I} or \ex{you} as the subject, as well as questions that mention the possessive pronouns \ex{my} or \ex{your}. 




\paragraph{Lack of Valid Questions.}
For some responses, no valid questions are generated -- i.e. all generated questions fail to pass the above filtering process. We use our NLI model as a fallback in such cases by taking its end-to-end prediction with $k$ as the hypothesis and $r$ as the premise. We set the score to be $1$ in case it predicts entailment, $0$ for contradiction, and 0.5 for the neutral case.



    



\begin{table*}[!ht]
    \centering
    \small
    \resizebox{\textwidth}{!}{
    \begin{tabular}{|p{0.08\linewidth}|p{0.48\linewidth}|p{0.55\linewidth}|}
    \hline
     \textbf{Topic} & \textbf{Response} & \textbf{Knowledge} \\
    \hline
    Coffee & coffee is \textcolor{red}{very acidic}. it has stimulating effects on humans. & Coffee is \textcolor{blue}{slightly acidic} and has a stimulating effect on humans because of its caffeine content. \\
    \hline
     French cuisine & in that time \textcolor{red}{italian cuisine was influenced by french cuisine} & During that time, \textcolor{blue}{French cuisine was heavily influenced by Italian cuisine}. \\
    \hline
    Madonna & she was born in \textcolor{red}{1968} and \textcolor{red}{raised in new york city.} & Born and \textcolor{blue}{raised in Michigan}, Madonna moved to New York City in 1978 to pursue a career in modern dance. \\
    \hline
    Sephora & me too! it's an \textcolor{red}{american fashion company founded in 1854}. & Sephora is a \textcolor{blue}{French chain of cosmetics stores founded in 1969}. \\
    \hline
    \end{tabular}
    }
    \caption{Examples for factually inconsistent responses from our dataset. Factual inconsistencies are marked in red, with their corresponding parts in the knowledge marked in blue. The first two examples are outputs of the \textit{dodeca}Dialogue system, and the last two are outputs of MemNet.}
    \label{tab:examples}
\end{table*}
\isection{Evaluation Benchmarks}{sec:tasks}

\isubsection{{W}izard of {W}ikipedia}{sec:wow}

The Wizard of Wikipedia dataset \citep[\wow;][]{dinan2019wizard} contains dialogues in which a bot needs to respond to user inputs in a knowledgeable way.  Each response should be grounded on a sentence from Wikipedia that is relevant to the conversation topic. Since this dataset does not contain explicit annotations for factual consistency of dialog responses, we construct a new dataset with such annotations for dialogues based on the \wow{} dataset as detailed in Section~\ref{sec:example_extraction}. 

\isubsection{Topical-Chat}{sec:topical}



Topical-Chat \citep{Gopalakrishnan2019} is a human-human knowledge-grounded conversation dataset. 
Each dialogue is accompanied by relevant Wikipedia pages, Washington Post articles and fun-facts from Reddit.
\citet{mehri-eskenazi-2020-usr} introduced USR, an evaluation metric that measures different aspects required from dialogue systems. 
To test USR, they collected human annotations on four different system responses and two human-generated responses for 60 dialog contexts from Topical-Chat. 
Each response was scored on a ``Uses Knowledge'' category, among others. Since a model that properly uses the knowledge is expected to use it in a factually consistent manner, we find it interesting to measure $Q^{2}$'s correlation with the human judgements for this category.

\subsection{Dialogue NLI}\label{sec:dnli}
Dialogue NLI  \citep[DNLI;][]{welleck-etal-2019-dialogue} is a dataset based on the Persona-Chat dialogue task \citep{zhang-etal-2018-personalizing}. It consists of pairs including either a personality description sentence or an utterance from the dialogue history (the \textit{premise}) and a subsequent dialogue utterance (the \textit{hypothesis}). Each pair is labeled as entailing, neutral, or contradicting.
A contradiction may be a clear logical contradiction, e.g. ``\emph{I have a dog}'' vs. ``\emph{I do not have a dog}'', but can also be two utterances that are not likely to be said by the same persona although they are not strict logical inconsistencies, e.g. ``\emph{i’m a manager}'' vs.``\emph{i’m a doctor}''.
Using this dataset, we test whether $Q^2$ can measure consistency when the grounding ``knowledge'' is a persona sentence or the previous dialogue history. 

\begin{table*}[t]
    \centering
    \resizebox{\textwidth}{!}{
    \begin{tabular}{|c|c|c|c|c|c|c|c|c|c|c|}
    \hline
    \textbf{system} & \textbf{data} & \textbf{\# questions} & \textbf{$Q^2$} &  \textbf{$Q^2$ w/o NLI} & \textbf{\% no answer} & \textbf{E2E NLI} & \textbf{Overlap($r$, $k$)} & \textbf{BLEU} & \textbf{BERTScore} \\
    \hline
     \multirow{3}{3.5em}{\textit{dodeca}} & Inconsistent & 328 & 0.238 & 0.159 & 54.88\% & 0.5 & 0.299 & 3.355 & 0.179 \\
    & Consistent & 341 & 0.696 & 0.516 & 15.25\% & 0.723 & 0.426 & 5.136 & 0.291 \\
    & Random sample & 258 & 0.496 & 0.349  & 29.84\% & 0.573 & 0.325 & 3.788 & 0.164 \\
    \hline
    \multirow{3}{3.5em}{MemNet} & Inconsistent & 324 & 0.135  & 0.123 & 62.04\% & 0.37 & 0.270 & 7.490 & 0.145 \\
    &  Consistent & 352 & 0.756  & 0.661 & 9.94\% & 0.717 & 0.526 & 20.145 & 0.376  \\
    & Random sample & 268 & 0.448 & 0.387 & 32.09\% & 0.537 & 0.337 & 11.654 & 0.183 \\
    \hline
    
    \end{tabular}
    }
    \caption{$Q^{2}$ and baseline scores on the annotated system responses from \wow{}.}
    \label{tab:wow_res}
\end{table*}

\section{Dataset Creation and Annotation}\label{sec:example_extraction}

To directly evaluate $Q^2$, we need an annotated dataset of knowledge-grounded dialogue responses and their factual consistency with respect to a given knowledge.
To obtain this, three of the paper's authors annotated the factual consistency of a random sample of responses from the following dialogue systems on the \wow{} validation set:
(1) \textbf{MemNet}, which is the model suggested by \citet{dinan2019wizard} for \wow{}.
(2) \textbf{\textit{dodeca}Dialogue},
which is the multi-task model fine-tuned on \wow{} in the \textit{dodeca}Dialogue benchmark \citep{shuster-etal-2020-dialogue}, as available in ParlAI\footnote{\url{https://parl.ai/docs/zoo.html}}~\citep{miller2017parlai}.
For both systems, we used beam search decoding with a beam size of 10, a beam block size of 3 and a context block size of 3 to generate responses. 

The annotators went through the responses until 150 examples of factually inconsistent responses were annotated for each system (300 in total), and then repeated the process and annotated the same number of factually consistent responses. 
The annotators
skipped factually consistent responses containing only general chit-chat with no reference to the grounding knowledge, such as ``\emph{Hi, how are you?}''. For factually inconsistent responses, they selected challenging examples in which the text seemed clear and coherent.
For each of the 600 extracted sentences, the annotation was extended to cover the outputs of both systems, resulting in 544 dialogue contexts and 1,088 annotated responses (due to overlaps). Out of the 544 contexts, 186 (34.2\%) were marked as inconsistent in the \textit{dodeca}Dialogue system and 274 (50.36\%) in the MemNet system.
The number of dialogue contexts and responses collected is comparable with those of other recently published datasets for dialogue evaluation, such as in \citet{mehri-eskenazi-2020-usr, pang-etal-2020-towards, zhao-etal-2020-designing}.

To evaluate the quality of the constructed dataset, 100 responses were sampled, and each annotator labeled them as consistent or inconsistent. The agreement level between annotators, measured by Fleiss' kappa, resulted in 0.853, representing high inter-annotator agreement.
Table~\ref{tab:examples} shows factually inconsistent responses from this dataset. 
Detecting some of these inconsistencies requires identifying subtle semantic divergences from the facts expressed by the knowledge.

\section{Experiments and Results}\label{sec:experiments}

To evaluate $Q^2$ as a metric we performed the following experiments for each dataset.





\isubsection{Wizard of Wikipedia}{sec:res_wow}


\paragraph{Absolute Scores.} Table~\ref{tab:wow_res} presents the $Q^2$ score for the different sets of annotated system responses, as well as for $150$ randomly selected system responses. We additionally report the total number of generated questions (\textit{after} filtering) for each set and the percentage of generated questions that had no answer in the knowledge. 
We denote our metric score by ``$Q^{2}$'', while ``$Q^{2}$ w/o NLI'' is an ablated variant obtained by dropping the NLI component and using the fallback token-level F1 instead, similarly to \citet{wang-etal-2020-asking}.

As we would expect from a metric measuring factual consistency of generative dialogue systems, the $Q^2$ score is indeed always highest for the consistent outputs, lowest for the inconsistent outputs, and in-between for random samples.
Assessing answer similarity using NLI results in higher absolute scores for both inconsistent and consistent responses, and by a larger margin for the latter.

\paragraph{Baselines.}
As baseline metrics, we first take the F1 token-level overlap of $r$ with $k$ as done in \wow{} \citep{dinan2019wizard}. We also use BLEU and BERTScore \citep{Zhang2020BERTScore} with the response $r$ as the output, and the knowledge $k$ as the reference. As our last baseline we run the NLI model described in \S\ref{sec:pipeline} in an end-to-end manner, taking $k$ as the premise and $r$ as the hypothesis. We set the score to be $1$ for the case of entailment and $0$ for contradiction. In the neutral case, we set the score to be $0.5$. The exact same settings are used as a fallback for $Q^{2}$ when no valid questions are generated.
As Table \ref{tab:wow_res} shows, the scores for the consistent data are higher than the scores for the inconsistent data for all baselines. However, in most cases, the score differences between the inconsistent data and the random samples are small, indicating that $Q^{2}$ better separates general responses from inconsistent ones.

\paragraph{Response-Level Evaluation.}
\begin{figure}[ht!]
\centering
 \includegraphics[width=1.0\linewidth]{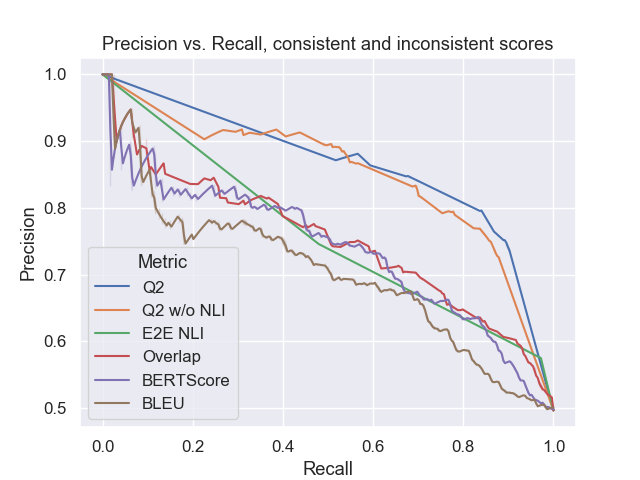}
 \caption{Precision-Recall curves for different response level score thresholds, calculated using the \textit{dodeca} and MemNet consistent and inconsistent examples.}
 \label{fig:precision_recall}
\end{figure}
To find if $Q^2$ can be used to automatically separate between consistent and inconsistent responses at the more granular, single response level, we report in Figure \ref{fig:precision_recall} the Precision/Recall curve of consistent responses for various response-level score thresholds for each evaluated metric on the \wow{} annotated data.

As Figure \ref{fig:precision_recall} shows, both $Q^2$ variants obtain higher precision and recall in comparison to the other metrics throughout the threshold values, suggesting that $Q^2$ is better at automatically separating between consistent and inconsistent examples at the response level. 
We additionally report in Table \ref{tab:prec-recall} the consistent and inconsistent Precision and Recall values for a threshold of 0.5. Responses with a score of 0.5 or below are classified as inconsistent and vice versa. The accuracy of the binary decision using this threshold is 77.3\% for $Q^{2}$, 73.1\% for $Q^{2}$ without the NLI-based answer spans comparison, and 65.3\% for the end-to-end NLI. We note that the threshold was arbitrarily selected for the purpose of demonstrating $Q^{2}$'s ability in separating consistent from inconsistent content, and properly tuning it by splitting the data into development and test sets may improve the results further.

\begin{table}[t]
    \centering
      \resizebox{\columnwidth}{!}{
    \begin{tabular}{|c|c|c|c|c|}
    \hline
     \textbf{Data split}  & \textbf{Metric} & \textbf{Precision} & \textbf{Recall} & \textbf{F1} \\
    \hline
     \multirow{3}{*}{Inconsistent} & $Q^{2}$ & \textbf{73\%} & 86.7\% & \textbf{0.793}\\
     & $Q^{2}$ w/o NLI & 67.1\% & \textbf{91\%} &  0.772 \\
     & E2E NLI & 61.2\% & 83.7\% & 0.707\\
    \hline  
     \multirow{3}{*}{Consistent} &  $Q^{2}$ & 83.5\% & \textbf{67.9\%} & \textbf{0.749}\\
     & $Q^{2}$ w/o NLI & \textbf{85.9\%} & 55.2\% & 0.672\\
     & E2E NLI & 74.1\% & 46.8\% & 0.574 \\
    \hline
    \end{tabular}
    }
    \caption{Precision-Recall values for consistent and inconsistent response detection, using a threshold of 0.5 for the binary decision.}
    \label{tab:prec-recall}
\end{table}

\paragraph{System-Level Evaluation.} We measure the correlation of each metric with human judgments for systems with varying inconsistency levels. To simulate such systems, we follow the method of \citet{graham-liu-2016-achieving} for MT evaluation. We first take dialogue contexts for which we have both a consistent and an inconsistent response, leaving us with 244 dialogue contexts (and 488 responses). We then bootstrap \citep{Efron1987BetterBC} by sampling 350 contexts (with repetition) for each simulated system $i$, ensuring that each system output contains $c_i\%$ factually inconsistent responses. Finally, we compute the system-level score for each system and the correlation between those scores and the human annotations. We repeat this $1000$ times and report average correlation and confidence intervals for each metric.

We take $c \in [0.05, 0.1, 0.15, 0.2, 0.25]$ as inconsistent response proportions for the simulated systems, and measure the Spearman correlation of $Q^{2}$ and the four baseline metrics with the human judgment scores of each system.
The results are detailed in Table \ref{tab:sys_level}. $Q^2$ obtains an average correlation of 0.9798, while the end-to-end NLI baseline, overlap, BERTScore, and BLEU obtain lower correlations of 0.9216, 0.878, 0.8467 and 0.3051, respectively. This suggests that $Q^{2}$ is better in evaluating factual consistency at the system-level.

\begin{table}[t]
    \centering
      \resizebox{\columnwidth}{!}{
    \begin{tabular}{|c|c|c|c|}
    \hline
      & \textbf{Avg. Correlation} & \textbf{Lower CI} & \textbf{Upper CI} \\
    \hline
    $Q^{2}$& \textbf{0.9798} & 0.9 & 1  \\
    \hline
     $Q^{2}$ w/o NLI & 0.9711 & 0.9 & 1  \\
    \hline
    E2E NLI  & 0.9216 & 0.6669 & 1 \\
    \hline
    Overlap($r$, $k$) & 0.878 & 0.5 & 1  \\
    \hline
    BERTScore & 0.8467 & 0.4 & 1  \\
    \hline
    BLEU & 0.3051 & -0.7 & 1  \\
    \hline
    \end{tabular}
    }
    \caption{Results for system level evaluation, taking systems with varying degrees of inconsistent outputs, and measuring the correlation between each system-level score and the human judgements.}
    \label{tab:sys_level}
\end{table}

\subsection{Topical-Chat}\label{sec:usr}
\citet{mehri-eskenazi-2020-usr} evaluated the correlation of their suggested metric, USR, as well as other existing automatic metrics, against human judgments on the Topical-Chat dataset \citep{Gopalakrishnan2019}. We note that in $8$ out of the $60$ examined dialogue contexts, no knowledge was used (the original dataset contains a "no fact" option). We thus experimented only with the $52$ knowledge-grounded dialogue contexts. We follow the settings of \citet{mehri-eskenazi-2020-usr}, which used only $5$ responses (out of the $6$ annotated per response), leaving out the original human response that was collected by \citet{Gopalakrishnan2019}. Accordingly, we are left with $260$ responses. Table~\ref{tab:usr} presents their reported correlation results for the ``Uses Knowledge'' category, as well as the correlation of $Q^2$ with the same human judgments. $Q^{2}$ demonstrates an improvement in this category that is statistically significant with $p < 0.001$ compared to the baselines. The contribution of the NLI component on this dataset resulted in even higher gains in terms of correlation in comparison to the \wow{} experiments, again showing the benefit of using our more intricate span comparison method.


\begin{table}[t]
    \centering
     \begin{small}
    \begin{tabular}{|c|c|c|}
    \hline
     \textbf{Metric} & \textbf{Spearman} & \textbf{Pearson} \\
    \hline
        $Q^{2}$ & \textbf{0.4579} & \textbf{0.4698}  \\
    \hline
        $Q^{2}$ w/o NLI & 0.3933 & 0.4105  \\
    \hline
        USR (best) & 0.4468 & 0.3175  \\
    \hline
     METEOR  & 0.3909 & 0.3328 \\
    \hline
    \end{tabular}
    \end{small}
    \caption{Correlation with human judgments for the ``Uses Knowledge'' category for different metrics. ``USR (best)'' stands for the best result achieved by \citet{mehri-eskenazi-2020-usr} for each category.}
    \label{tab:usr}
\end{table}

\subsection{Dialogue NLI}\label{sec:dnli-res}

We test $Q^2$'s applicability for measuring persona consistency and self-consistency between dialogue utterances, as described in \S\ref{sec:dnli}. We calculate the $Q^2$ score for each persona-utterance or utterance-utterance pair and choose a threshold of 0.1 for predicting entailment or contradiction by tuning on the development set. Since a dialogue utterance should be grounded in the personality description or in the conversation's history, we treat neutral claims as inconsistent, and expect $Q^2$ to address them as contradictions. 
As DNLI aims at testing persona consistency, we avoid filtering out questions that include personal or possessive pronouns. 

\begin{table}[t]
    \centering
     \begin{small}
    \begin{tabular}{|c|c|}
    \hline
\textbf{Model} &\textbf{Accuracy} 
\\\hline
$Q^2$  &\textbf{74.49}\% \\\hline
Baseline -- NLI only &67.42\% \\\hline
InferSent SNLI &47.03\% \\ \hline
InferSent Hyp. Only &51.52\%  \\ \hline
\end{tabular}
\end{small}
\caption{Accuracy on the DNLI dataset, Test Gold.}
\label{tab:dnli}
\end{table}


Table~\ref{tab:dnli} presents $Q^2$'s accuracy on the Test Gold split of DNLI, compared to other zero-shot methods.  
Our first baseline uses the NLI model in $Q^{2}$ in the end-to-end manner described above (``Baseline -- NLI only''), which is similar to the approach of \citet{welleck-etal-2019-dialogue,pang-etal-2020-towards}. To be comparable with $Q^2$'s binary decision, we allow neutral claims to be predicted as either neutral or contradicting.
We also show results from zero-shot methods reported in \citet{welleck-etal-2019-dialogue}: a model that uses the hypothesis sentence only (``InferSent Hyp. Only'') and a model trained on the SNLI dataset but evaluated on DNLI (``InferSent SNLI'').
$Q^{2}$ performs better than the end-to-end NLI baselines, indicating that our QG/QA approach with NLI is more robust than simply applying end-to-end NLI with full sentences or passages. 

\subsection{Analysis}\label{sec:analysis}
The results on the three datasets demonstrate $Q^{2}$'s zero-shot, reference-response-free capability to generalize to various dialogue tasks that require evaluation of factual consistency.
To shed more light on our approach we performed the following qualitative and quantitative analyses.

\paragraph{Robustness to Underlying Model Quality.} 
The performance of $Q^{2}$ depends on the different components used throughout the pipeline, i.e., the QG, QA, and NLI models. To demonstrate that $Q^{2}$ is robust to the quality of these models, we experiment with using smaller models in the pipeline.
First, we replace the T5-base model for question generation with a T5-small model, again fine-tuned on SQuAD1.1. Next, we replace the Albert-Xlarge QA model with Albert-base, similarly fine-tuned on SQuAD2.0 for question answering. 

As Table \ref{tab:sys_level_ablations} shows, the correlations with human judgments are barely influenced by using smaller QG/QA models, showing the robustness of our method to changes in the underlying models. Table \ref{tab:pipeline_robustness} presents the absolute scores of the smaller models on the \wow{} dataset, as well as  each variant's question coverage, defined as the percentage of responses for which $Q^{2}$ generated at least one valid question, not resorting to the end-to-end NLI fallback. While the question coverage slightly decreases when using smaller models, the gap between consistent and inconsistent scores remains unaffected. As we expected, a smaller QG model results in lower $Q^{2}$ scores, for all data splits. Surprisingly, using a smaller QA model had the opposite outcome - higher $Q^{2}$ scores in all cases.

Regarding domain robustness of the undelying models, while the QG and QA models were trained on a dataset collected from Wikipedia and are therefore suited for \wow{}'s domain, these models work well even when the grounding source is not Wikipedia. This is the case in Topical-Chat, in which each dialogue is accompanied by Washington Post articles and fun-facts from Reddit in addition to pages from Wikipedia; and in the DNLI dataset, which deals with persona and self-consistency of dialogue systems and does not contain any references to Wikipedia.

\begin{table}[!t]
    \centering
      \resizebox{\columnwidth}{!}{
    \begin{tabular}{|c|c|c|c|}
    \hline
      & \textbf{Avg. Correlation} & \textbf{Lower CI} & \textbf{Upper CI} \\
    \hline
    Original $Q^{2}$& \textbf{0.9798} & 0.9 & 1  \\
    \hline
    T5-small & 0.9722 & 0.9 & 1  \\
    \hline
    Albert-base & 0.9797 & 0.9 & 1  \\
    \hline
    \end{tabular}
    }
    \caption{Correlations with human judgements when using a smaller QG or a smaller QA model.}
    \label{tab:sys_level_ablations}
\end{table}

\begin{table}[t]
    \centering
    \resizebox{\columnwidth}{!}{
    \begin{tabular}{|c|c|c|c|c|c|}
    \hline
    \textbf{Data} & \textbf{Model} & \textbf{Coverage} & \textbf{$Q^2$} & \textbf{$Q^2$ w/o NLI}  \\
    \hline
     \multirow{3}{5.5em}{\textit{dodeca} inconsistent} & Original & 92.67\% & 0.238 & 0.159 \\
    & T5-small & 90.67\% & 0.198  & 0.143 \\
    & Albert-base & 92\% & 0.293  & 0.213 \\
    \hline
     \multirow{3}{5.5em}{\textit{dodeca} consistent} & Original & 94\% & 0.696 & 0.516 \\
    & T5-small & 90.67\% & 0.601 & 0.44 \\
    & Albert-base & 92.67\% & 0.709 & 0.534  \\
    \hline
     \multirow{3}{5.5em}{MemNet inconsistent} & Original & 94.67\% & 0.135 & 0.123 \\
    & T5-small & 90\% & 0.104 & 0.099 \\
    & Albert-base & 94\% & 0.189 & 0.134 \\
    \hline
    \multirow{3}{5.5em}{MemNet consistent} & Original & 92.67\% & 0.756  & 0.661 \\
    & T5-small & 88.67\% & 0.705 & 0.613 \\
    & Albert-base & 89.33\% & 0.791 & 0.7 \\
    \hline
    
    \end{tabular}
    }
    \caption{$Q^{2}$'s results on \wow{} when using a smaller QG or a smaller QA model. Coverage refers to the questions coverage, i.e., the percentage of responses for which $Q^{2}$ generated at least one valid question.}
    \label{tab:pipeline_robustness}
\end{table}

\paragraph{Lack of Valid Questions.}
For some responses, $Q^2$ does not generate any valid questions. When testing the extent of this phenomenon in the inconsistent vs. consistent samples collected based on the MemNet and \textit{dodeca}Dialogue outputs, a similar proportion of around $6$-$8\%$ responses had no valid questions. The proportion of such responses in the randomly sampled examples is much higher -- around $20\%$. As mentioned in \S\ref{sec:pipeline}, we handle such cases using an end-to-end NLI fallback.

The higher proportion of such responses in the random samples indicates that lack of valid questions is more common in general chit-chat than in knowledge-grounded content. This raises the need to improve the identification and separation of general chit-chat responses from more ''knowledgable'' ones, which we plan to address in future work.

Another cause for low-quality questions that do not pass the filtering process is responses that contain pronouns referring to entities in the dialogue history  -- e.g. ``\textit{he won an album of his own in 2015}'' requires resolving ``\textit{he}''. Preliminary experiments with adding a coreference resolution step to our pipeline showed increased coverage, and we plan to further address this gap in future work.


\paragraph{Qualitative Analysis.}
To get a better impression of $Q^2$'s operation, we give examples of how it operates in its various stages.
Figure \ref{fig:pipeline} presents an example for an inconsistent response, together with a generated question and the answer $Q^2$ obtained based on the knowledge. In this example, the question was unanswerable using the knowledge, thus the score for this question is $0$. Indeed, this is the desired score, as the knowledge didn't mention that coffee is \textit{very} acidic.

Another example for successful output is for the following response: \ex{i'm not sure about that but i do know that they are reliant on vulnerable species!}, generated by the \textit{dodeca}Dialogue system when conversing about giant Pandas, while conditioning on the following knowledge paragraph: \ex{The giant panda is a conservation reliant vulnerable species.}. The response is clearly inconsistent with the knowledge as Pandas are reliant on conservation and not on vulnerable species. Here, $Q^{2}$ extracted \ex{vulnerable species} as an informative span, and generated the question: \ex{What are they reliant on?}. The answer to this question using the knowledge was \textit{``conservation''}, which resulted in assigning this question a score of $0$.

These examples also demonstrate a major advantage of $Q^{2}$, being self-explanatory and interpretable. Other than the final score, $Q^{2}$ outputs the generated questions, the response-based answer spans and the answers the QA model predicted based on the knowledge, which can be used as an explanation to the assigned score or to highlight the potentially inconsistent text spans in the response.


Some errors of $Q^{2}$ are caused by generating questions for the chit-chat parts of responses. In a conversation regarding the color purple, the \textit{dodeca}Dialogue system generated the response: \ex{purple is my favorite color. it's between red and blue.}, when the knowledge was: \ex{Purple is a color intermediate between blue and red.} Even though the response used the knowledge faithfully, one out of two valid generated questions for it was \ex{What is purple?}, for which the response-based answer is \ex{my favorite color}, while the knowledge-based answer is, of course, different. 

\isection{Related Work}{sec:related}

\paragraph{Automatic Evaluation of Dialogue Systems.}
Automatically evaluating natural language generation is a notoriously difficult problem, especially when considering open-ended tasks such as dialogue. Standard token-matching metrics, such as BLEU~\citep{papineni-etal-2002-bleu} or METEOR~\citep{banerjee-lavie-2005-meteor} in machine translation, or ROUGE~\citep{lin-2004-rouge} in summarization, were shown to have weak or no correlation with human judgements for dialogue~\citep{liu-etal-2016-evaluate, lowe-etal-2017-towards}. Supervised assessment methods learn to predict human-like evaluation scores~\citep{lowe-etal-2017-towards}, but they require a significant annotation effort for achieving training data.
Recently, \citet{mehri-eskenazi-2020-usr} and \citet{pang-etal-2020-towards} suggested to use large pretrained language models~\citep{liu2019roberta, radford2019language} to develop reference-response-free metrics for dialogue evaluation. Such LMs are also the backbone of the QG, QA and NLI models employed in $Q^2$.

\paragraph{Factual Consistency and Hallucinations.}
Factual consistency in summarization has attracted increasing attention in recent years~\citep{maynez-etal-2020-faithfulness} both in improving factual consistency of abstractive summarization systems~\citep{Cao2018FaithfulTT} and in evaluating the factual consistency of generated summaries~\citep{goodrich2019assessing, kryscinskiFactCC2019,xu-etal-2020-fact}.
Factual inconsistency
has been observed in neural machine translation \citep{lee2019hallucinations} mainly when considering out-of-domain scenarios~\citep{koehn-knowles-2017-six, wang-sennrich-2020-exposure, muller-etal-2020-domain}.

Concurrently with our work, \citet{dziri2021evaluating} introduced the Benchmark for Evaluation of Grounded INteraction (BEGIN). BEGIN consists of \wow{}-based dialogue turns annotated for factual consistency with respect to the grounding knowledge. BEGIN models the task of evaluating groundedness as an NLI task and examples are annotated with five labels: entailment, contradiction, hallucination, off-topic and generic, where the last three are all considered to be neutral from an NLI perspective. Also relevant to our work, \citet{rashkin-etal-2021-increasing} showed that faithfulness in knowledge-grounded dialogues can be improved by using controllable features based on NLI model predictions.


\paragraph{Evaluation via Question Answering and Question Generation.}
QA-based evaluation metrics have been proposed as a means for measuring content coverage in text generation tasks. For example, \citet{eyal-etal-2019-question} used QA models for abstractive summarization both as an evaluation metric and as an optimization criterion that improved the downstream ROUGE scores by manually constructing questions around entities in the source document. These metrics aim at assessing whether key information from the input documents is expressed in the summaries (Recall-oriented). \citet{durmus-etal-2020-feqa} and \citet{wang-etal-2020-asking} suggested using QG and QA to identify factual inconsistencies in abstractive summaries, which is more Precision-oriented. Their approach is based on the intuition that if a summary is consistent with its source, questions asked on the summary and the source should result in similar answers. Recently, \citet{scialom2020QuestEval} suggested QuestEval, which combines the Recall and Precision oriented QG and QA approaches, obtaining a more robust metric for evaluating abstractive summaries which was adopted in the GEM shared task \citep{gem-2021-natural}. To overcome the low scores assigned by the token-level F1 measure to semantically-identical answers that are lexically different, they use a measure of the QA confidence of answerability \citep{scialom-etal-2019-answers}, which is the complement of the probability that the QA model gives to the ``no answer'' prediction.  This measure reflects the answerability independently of the way the answer is expressed, but does not take into account possible model hallucinations, and it is therefore only applied for the Recall-based component. Our suggested NLI-based answer comparison allows lexical variability in the Precision-based component as well.

Comparing to other automatic evaluation methods of abstractive summaries, the QG-QA based methods showed higher correlations with human judgments of factual consistency. To the best of our knowledge, our work is the first to apply a QG-QA approach for evaluating dialogue generation.

\section{Conclusion and Future Work}
We presented $Q^2$, an automatic evaluation method for factual consistency in knowledge grounded dialogue. $Q^2$ employs question generation, question answering and NLI models, and does not require reference responses.
To test our approach, we compiled a dataset of dialogue responses from two systems on the Wizard of Wikipedia dataset, which we annotated for factual consistency. Extensive experiments on this dataset, as well as on the Topical-Chat and DialogueNLI datasets, present strong results for $Q^2$ against various baselines. In future work we would like to map parts of a response to different types like chit-chat, persona and factual, in order to evaluate each against its appropriate source of truth. 
Other directions for future research are to apply $Q^2$ in additional tasks where factual consistency is essential, such as automated fact-checking \cite{thorne-vlachos-2018-automated}, and to use its evaluation signal to improve the factual consistency of generation models as proposed by \citet{rashkin-etal-2021-increasing} or \citet{nan-etal-2021-improving}.

\section*{Acknowledgements}

This work was carried out as part of a Master Sponsored Research Agreement between the Hebrew University and Google, and was also supported by a gift from Google. Or Honovich was partially supported by a fellowship from the Hebrew University Center for Interdisciplinary Data Science Research.

\bibliography{anthology,custom}
\bibliographystyle{acl_natbib}

\clearpage

\input{appendix.tex}

\end{document}

%% file: appendix.tex
\appendix

\begin{table}[t]
    \centering
    \resizebox{\columnwidth}{!}{
    \begin{tabular}{|c|c|c|c|c|}
    \hline
    \textbf{Data} & \textbf{Model} & \textbf{Coverage} & \textbf{Score} \\
    \hline
     \multirow{3}{5.5em}{\textit{dodeca} inconsistent} & $Q^2$ & $92.67\%$ & $0.238$  \\
    & -top-n & $87.33\%$ & $0.265$  \\
    & -filter personal & $92.67\%$ & $0.243$   \\
    \hline
     \multirow{3}{5.5em}{\textit{dodeca} consistent} & $Q^2$ & $94\%$ & $0.696$ \\
    & -top-n & $85.33\%$ & $0.7$   \\
    & -filter personal & $90\%$ & $0.675$  \\
    \hline
     \multirow{3}{5.5em}{MemNet inconsistent} & $Q^2$ & $94.67\%$ & $0.135$  \\
    & -top-n & $84.67\%$ & $0.153$  \\
    & -filter personal & $86\%$ & $0.139$  \\
    \hline
    \multirow{3}{5.5em}{MemNet consistent} & $Q^2$ & $92.67\%$ & $0.756$ \\
    & -top-n & $85.33\%$ & $0.729$  \\
    & -filter personal & $88\%$ & $0.719$  \\
    \hline
    
    \end{tabular}
    }
    \caption{Results for the ablations studies.}
    \label{tab:additional_experiments}
\end{table}

\section{Ablation Study}\label{sec:ablation}
Table \ref{tab:additional_experiments} presents the results of two ablations studies on $Q^{2}$. We show the scores obtained in these studies, as well as the question coverage, defined as the percentage of responses for which $Q^{2}$ generated at least one valid question, not resorting to the end-to-end NLI fallback.

First, we experiment with a different decoding strategy for generating questions. Instead of using beam search and taking the $n$ top-ranked generated questions (see \S\ref{sec:pipeline}), we use greedy decoding, generating only one question per answer candidate. Next, we additionally drop the filtration of questions relating to personal statements and opinionated parts of the response.   


\paragraph{Top-n Questions.}
Contrary to our expectations, When applying greedy decoding and taking a single question per an informative span, we inspect an increase for all data splits, except for the MemNet consistent responses. While the top-$n$ decoding seems to be ineffective in terms of separating consistent responses from inconsistent responses, it is effective for improving the question coverage of $Q^{2}$.

\paragraph{Filtering Questions Relating to Personal Statements.}
As mentioned in \S\ref{sec:pipeline}, we filter questions that ask about personal statements expressed by the model. Examples of such questions are ``What do I love?'', which was generated given the text ``I love cats'' and the informative span `cats'. Such text should not be evaluated for factual consistency and is allowed regardless of the knowledge. We report here the results for dropping this filtering step, on top of the previous experiment (applying greedy decoding). As Table \ref{tab:additional_experiments} shows, when not removing such questions, scores are lower for all data splits. Naturally, the question coverage increases.


\begin{table}[t]
    \centering
      \resizebox{\columnwidth}{!}{
    \begin{tabular}{|c|c|c|}
    \hline
      & \textbf{$Q^{2}$} & \textbf{\% no answer} \\
    \hline
     Same dialogue & $0.02$ & $91.02\%$  \\
    \hline
   Random dialogue& $0$ & $99.61\%$  \\
    \hline
    \end{tabular}
    }
    \caption{Results using randomly selected knowledge.}
    \label{tab:random_res}
\end{table}

\section{Computing Infrastructure}
We ran each experiment on 4 CPUs. For each data split (i.e., $150$ responses), the runtime was $\sim1.5-2$ hours. In future work, we plan to design a more efficient version of $Q^{2}$.

\section{Additional Experiments}\label{sec:additional}
\paragraph{Random Knowledge.}
We replace the knowledge $k$ with randomly selected knowledge to test the sensitivity of our method to such adversarial cases. Two variants of knowledge selection are applied: In the first variant, we randomly select knowledge from the same dialogue, but from a different turn. In the second, we randomly select knowledge from a different dialogue. In both cases, we expect $Q^{2}$'s score to be extremely low, as the knowledge should have little (in the first variant) to no (in the second variant) relation with $r$. 
Table~\ref{tab:random_res} shows the results for using randomly selected knowledge;
As expected, in both cases more than $91\%$ of the generated questions had no answer in the knowledge, and this is more severe ($99.6\%$) when using knowledge from a different dialogue.

\begin{table}[t]
    \centering
      \resizebox{\columnwidth}{!}{
    \begin{tabular}{|c|c|c|c|}
    \hline
      & \textbf{Average \# Characters} & \textbf{Average \# Tokens}  \\
    \hline
     Inconsistent & $70.84$ & $15.79$   \\
    \hline
    Consistent &$69.49$ & $15.13$   \\
    \hline
    Random & $69.44$ & $15.86$   \\
    \hline
    \end{tabular}
    }
    \caption{Average sentence length and average number of tokens per sentence in our collected dataset.}
    \label{tab:length}
\end{table}

\paragraph{Response Length.}
To test whether simple ``surface markers'' can differentiate consistent responses from inconsistent responses, we compare the average number of characters and the average number of tokens for responses in our dataset. As Table \ref{tab:length} shows, no strong differences were found for the \textit{dodeca} system outputs. Similar results were obtained for the MemNet system.

\section{Additional Graphs} \label{appendix_graph}
\begin{figure*}[h!]
    \begin{subfigure}{0.5\textwidth}
        \includegraphics[width=1\linewidth]{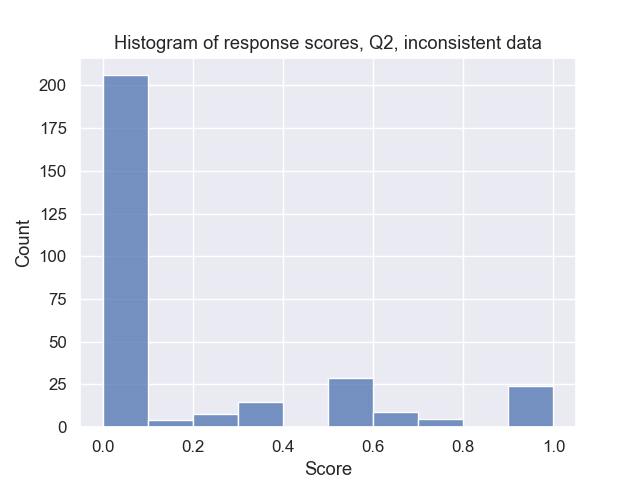}
        \caption{}
        \label{fig:hist_q_incons}
    \end{subfigure}
    \begin{subfigure}{0.5\textwidth}
        \includegraphics[width=1\linewidth]{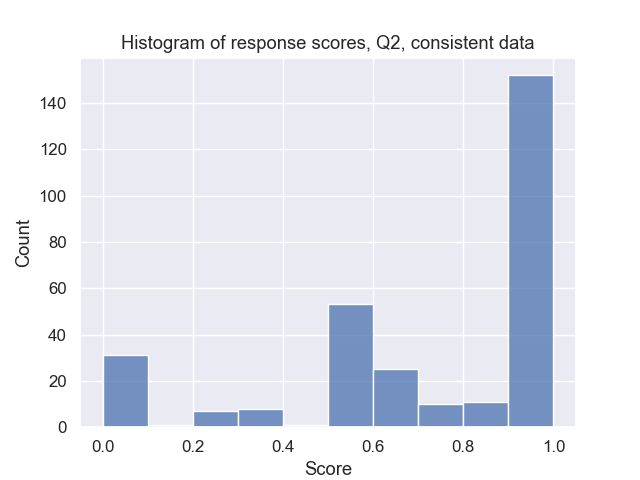}
        \caption{}
        \label{fig:hist_q_cons}
    \end{subfigure}
\caption{Distribution of the response-level scores for $Q^{2}$. (a) Distribution for the inconsistent data. (b) Distribution for the consistent data.}
\label{fig:hist_q_nli}
\end{figure*}

\begin{figure*}[h!]
    \begin{subfigure}{0.5\textwidth}
        \includegraphics[width=1\linewidth]{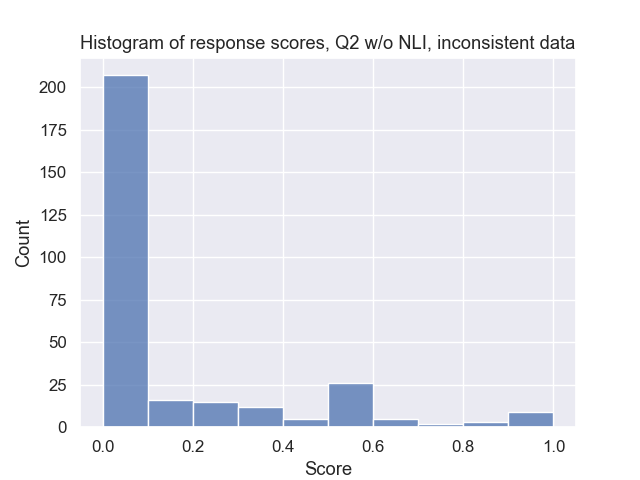}
        \caption{}
        \label{fig:hist_q_tok_incons}
    \end{subfigure}
    \begin{subfigure}{0.5\textwidth}
        \includegraphics[width=1\linewidth]{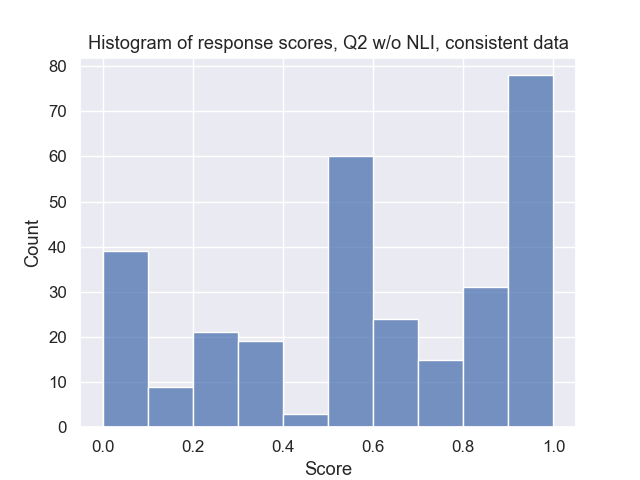}
        \caption{}
        \label{fig:hist_q_tok_cons}
    \end{subfigure}
\caption{Distribution of the response-level scores for $Q^{2}$ w. token-matching. (a) Distribution for the inconsistent data. (b) Distribution for the consistent data.}
\label{fig:hist_q}
\end{figure*}

\begin{figure*}[h!]
    \begin{subfigure}{0.5\textwidth}
        \includegraphics[width=1\linewidth]{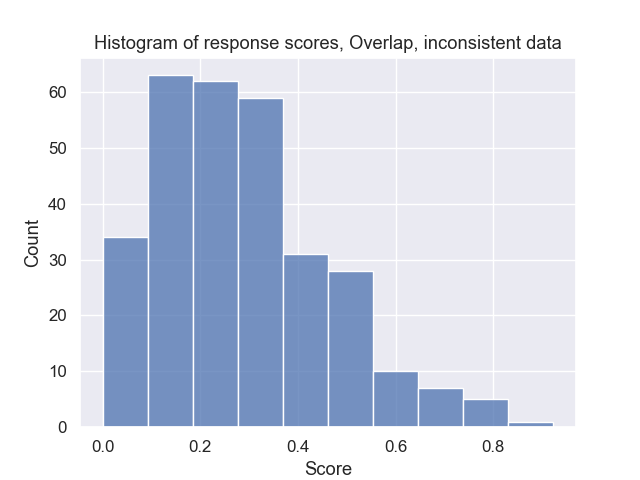}
        \caption{}
        \label{fig:hist_overlap_incons}
    \end{subfigure}
    \begin{subfigure}{0.5\textwidth}
        \includegraphics[width=1\linewidth]{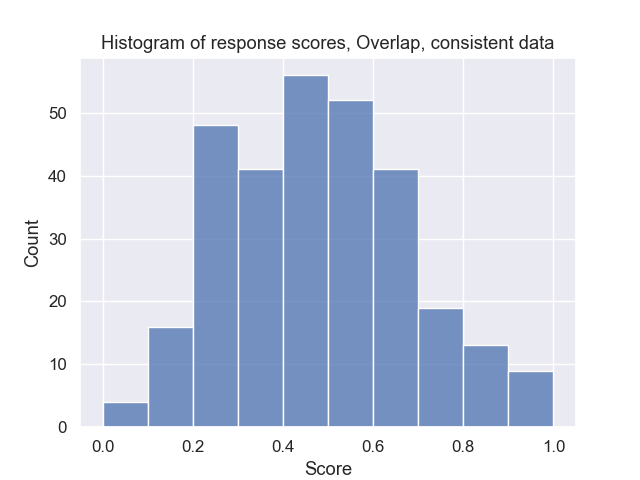}
        \caption{}
        \label{fig:hist_overlap_cons}
    \end{subfigure}
\caption{Distribution of the response-level scores for the overlap baseline. (a) Distribution for the inconsistent data. (b) Distribution for the consistent data.}
\label{fig:hist_overlap}
\end{figure*}

Figures \ref{fig:hist_q_nli} -- \ref{fig:hist_overlap} show the distribution of the response-level scores assigned by $Q^{2}$ and by the Overlap($r$, $k$) baseline for the consistent and inconsistent data.

\section{Annotation Guidelines}
\footnote{The guidelines are based on the insights provided by \citet{durmus-etal-2020-feqa} regarding annotating faithfulness.}
In this task, you will be presented with dialogues spanning various topics, conducted with a bot.

In each turn of the conversation, the bot was provided with a Wikipedia sentence relevant to the conversation topic and the current context of the conversation. The knowledge, or pieces of it, are integrated into the conversation.

\paragraph{Inconsistent responses collection}

You will be asked to detect bot responses that are \textit{inconsistent} with the given knowledge. Such inconsistencies may include:
\begin{enumerate}
    \item Information that was not at all mentioned by the knowledge.
    \item Changes to the knowledge, resulting in information that was not expressed by it. Note that these changes may be subtle.
\end{enumerate}
When marking a response as inconsistent, please:
\begin{enumerate}
    \item Check if the response is clear and coherent. If not, ignore the response. 
    \item Ignore your background knowledge and focus on the information provided to the bot.
\end{enumerate}

\paragraph{Consistent responses collection}

You will be asked to detect bot responses that are \textit{consistent} with the given knowledge.
When marking a response as consistent, please:
\begin{enumerate}
    \item Check if the response is clear and coherent. If not, ignore the response. 
    \item Select a response only if it uses the given knowledge. Ignore responses that are uninformative and only contain chit-chat.
\end{enumerate}

